\documentclass[runningheads]{llncs}
\usepackage{graphicx}

\usepackage{tikz}
\usepackage{comment}
\usepackage{amsmath,amssymb} 
\usepackage{color}
\usepackage{multirow}
\usepackage{subfigure}
\usepackage{algorithmicx}  
\usepackage{algpseudocode}
\usepackage{booktabs,tabularx}
\usepackage[ruled]{algorithm2e} 

\usepackage{verbatim}

\usepackage{setspace}
\usepackage{array}
\usepackage[misc,geometry]{ifsym}

\newcommand{\ie}{\textit{i}.{e}.}

\makeatletter
\def\hlinew#1{%
  \noalign{\ifnum0=`}\fi\hrule \@height #1 \futurelet
   \reserved@a\@xhline}
\makeatother

\newcolumntype{P}[1]{>{\raggedright\arraybackslash}p{#1}}
\newcolumntype{M}[1]{>{\centering\arraybackslash}m{#1}}

\definecolor{myGreen}{RGB}{55,149,73}

\begin{document}
\pagestyle{headings}
\mainmatter
\def\ECCVSubNumber{897}  

\title{Optimal Boxes: Boosting End-to-End Scene Text Recognition by Adjusting Annotated Bounding Boxes via Reinforcement Learning} 


\titlerunning{Optimal Boxes}
%

\author{
  Jingqun Tang \inst{1}* \and
  Wenming Qian\inst{2}* \and
  Luchuan Song\inst{3} \and
  Xiena Dong\inst{4} \and
  Lan Li \inst{5} \and
  Xiang Bai \inst{6}$^{\textrm{\Letter}}$ 
}

\authorrunning{Jingqun Tang, et al.}
%
\institute{
  Ant Group \\
  \email{jingquntang@163.com} \and
  NetEase Fuxi AI Lab \\
  \email{wenmingqian@corp.netease.com}\and
  University of Rochester \\
  \email{lsong11@ur.rochester.edu}\and
  Hangzhou Dianzi University\\
  \email{dxn@hdu.edu.cn} \and
  Wuhan University\\
  \email{2016302580090@whu.edu.cn}\and
  Huazhong University of Science and Technology \\
  \email{xbai@hust.edu.cn}
  }

\maketitle

\let\thefootnote\relax\footnotetext{$^*$ Equal contribution.  $^{\textrm{\Letter}}$ Corresponding author.}

\begin{abstract}
Text detection and recognition are essential components of a modern OCR system. Most OCR approaches attempt to obtain accurate bounding boxes of text at the detection stage, which is used as the input of the text recognition stage. We observe that when using tight text bounding boxes as input, a text recognizer frequently fails to achieve optimal performance due to the inconsistency between bounding boxes and deep representations of text recognition. In this paper, we propose Box Adjuster, a reinforcement learning-based method for adjusting the shape of each text bounding box to make it more compatible with text recognition models. Additionally, when dealing with cross-domain problems such as synthetic-to-real, the proposed method significantly reduces mismatches in domain distribution between the source and target domains. Experiments demonstrate that the performance of end-to-end text recognition systems can be improved when using the adjusted bounding boxes as the ground truths for training. Specifically, on several benchmark datasets for scene text understanding, the proposed method outperforms state-of-the-art text spotters by an average of 2.0\% F-Score on end-to-end text recognition tasks and 4.6\% F-Score on domain adaptation tasks.
\end{abstract}

\section{Introduction}
\begin{figure}[t]
\includegraphics[width=0.8\linewidth] {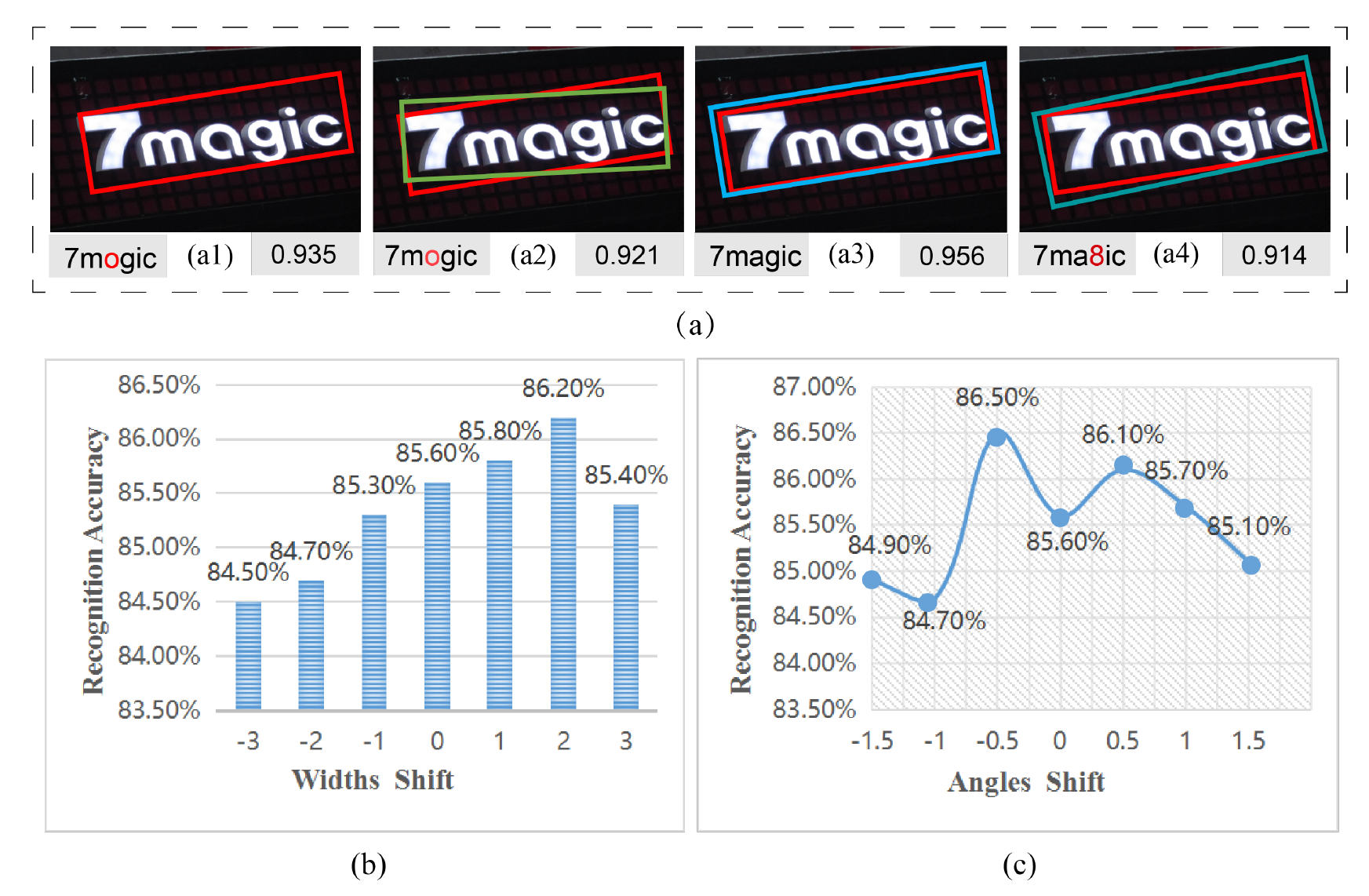}
  \centering
   \caption{(a) The red boxes represent the ground-truth bounding boxes, while the others are randomly shifted. Fig.(a1) represents the recognition confidence and recognition results with the ground-truth bounding box, while Fig.(a2) to Fig.(a4) with randomly shifted bounding boxes. The recognition results are presented on the left of (a1) to (a4), and the recognition confidence on the right; (b) text recognition accuracy with adjusting widths of the ground-truth bounding boxes; (c) text recognition accuracy with adjusting angles of the ground-truth bounding boxes.}
   \label{fig:intro}
\end{figure}


In modern society, text plays a more important role than ever before as an essential tool for communication and collaboration. Meanwhile, scene text reading has become an active research area due to its wide applications in the real world, such as image instant translation~\cite{2009Method,song2021fsft}, image search~\cite{2011Exploiting,2011Mobile}, and industrial automation~\cite{2013Extracting,2005A}.

Text detection and recognition can be roughly divided into two categories: two-step systems and end-to-end systems. For two-step systems~\cite{EAST,DB,CRAFT,tang2022few,crnn,rare,li2019show}, since detected texts are cropped from the image, detection and recognition are two separate steps. Some of these methods first generate text proposals using a text detection model and then recognize them with a text recognition model~\cite{2016Reading,liao2017textboxes,2016Synthetic}. For end-to-end systems, many end-to-end trainable networks~\cite{DBLP:conf/aaai/BartzYM18,2017Deep,he2018end,Li_2017_ICCV,2018FOTS} have recently been proposed. ~\cite{2017Deep,he2018end,2018FOTS} develop unified text detection and recognition systems with very similar overall architectures, which consist of a recognition branch and a detection branch. 
However, current models simply use tight annotated text bounding boxes as the ground truth, ignoring the inconsistency between bounding boxes and deep representations of text recognition. So, are tight bounding boxes the most suitable for recognition tasks? Through a series of experiments, we observe that a text recognizer frequently fails to achieve its best performance when using tight bounding boxes as inputs. As shown in Fig.\ref{fig:intro}(a), with suitable adjustments to the bounding boxes, we can get higher recognition confidence and correct recognition results (see Fig.\ref{fig:intro}(a3)). As shown in Fig.\ref{fig:intro}(b) and Fig.\ref{fig:intro}(c), the text recognizer can perform better when adjusting the widths or rotation angles of the ground-truth bounding boxes. The above experiments show a certain inconsistency between bounding boxes and deep representations of text recognition. Additionally, unlike in COCO~\cite{coco}, where clipping two pixels off an object does not prevent recognition, a 1-2 pixel error in text boxes may render the correct recognition prediction unrecoverable. The text recognition result is more sensitive to changes in the bounding box. To address the aforementioned problems, this paper presents a reinforcement learning-based method for adjusting the shape of each ground-truth bounding box so that it is more compatible with the text recognition task.

We propose a reinforcement learning-based method named Box Adjuster, which mitigates the inconsistency between bounding boxes and deep representations of text recognition. Our method can be summarized as follows: Firstly,  we choose a range of representative text recognizers and regard the average recognition confidence as a reward. Secondly, the \textbf{Box} Adjusting \textbf{D}eep \textbf{Q} \textbf{N}etwork (BoxDQN) with Feature Fusion Module (FFM) is trained, which can automatically adjust bounding boxes according to the text recognition reward. Finally, we train the end-to-end scene text recognition model with the refined ground-truth bounding boxes for better recognition.
Furthermore, as a preprocessing method, it is only applied in the process of creating training datasets. Thus, there is no additional computational cost in the forward phase.

Additionally, the proposed Box Adjuster is beneficial for resolving cross-domain problems such as synthetic-to-real, in which the source domain represents labeled synthetic data and the target domain represents unlabeled real data. To prove the effectiveness and generalization of our approach, we conduct experiments on standard benchmarks, including ICDAR 2013~\cite{ic13}, 
ICDAR 2015~\cite{ic15}, ICDAR 19-ReCTS~\cite{rects} and ICDAR 19-MLT~\cite{ic19mlt} datasets. The proposed method achieves better performance on the datasets when compared with the existing state-of-the-art methods. Besides, we demonstrate the efficacy of our approach on domain adaptation tasks.

Our contributions can be summarized as follows:
\begin{itemize}
\item We introduce the Box Adjuster, which adjusts the shape of each annotated text bounding box to make it more compatible with text recognition models. 
Besides, a text recognition-based reward is proposed to train our BoxDQN model in order to capture optimal annotated bounding boxes.
\item Our proposed Feature Fusion Module (FFM), which integrates foreground, background, and box coordinates, considerably enhances BoxDQN in terms of application scope and accuracy.
\item Our approach is generalized and can be easily applied to boost existing OCR systems without any additional computational cost during the inference phase. Concurrently, the proposed method outperforms state-of-the-art text spotters by an average of 2.0\% F-Score on public datasets.
\item When utilized in the cross-domain area, the proposed method significantly mitigates inconsistency between source and target domains, resulting in an average improvement of 4.6\% for state-of-the-art text spotters.
\end{itemize}

\section{Related Works}

\subsection{Two-Step OCR Systems}

In two-step systems, due to the fact that detected texts are cropped from the image, the detection and recognition are two separate steps. Some of these methods first generate text proposals using a text detection model~\cite{EAST,DB,tang2022few} and then recognize them with a text recognition model~\cite{2016Reading,liao2017textboxes,2016Synthetic}. Jaderberg et al.~\cite{2016Reading} use a combination of Edge Box proposals~\cite{2014Edge} and a trained aggregate channel features detector~\cite{2014Fast} to generate candidate text bounding boxes. Liao et al.~\cite{liao2017textboxes} combine an SSD~\cite{liu2016ssd} based text detector and CRNN~\cite{crnn} to spot text in images. In addition, for the detection step, EAST~\cite{EAST} further simplifies the anchor-based detection by adopting the U-shaped design~\cite{10.1007/978-3-319-24574-4_28} to integrate features from different levels. And for the recognition step, RARE~\cite{rare} consists of a Spatial Transformer Network (STN) and a Sequence Recognition Network (SRN), which is robust to irregular text. One major disadvantage of two-step methods is that the propagation of error between the recognition and detection models will result in less satisfactory performance.

\subsection{End-to-End OCR Systems}

Many end-to-end trainable networks have recently been proposed~\cite{DBLP:conf/aaai/BartzYM18,2017Deep,Li_2017_ICCV,he2018end,2018FOTS}. Bartz et al.~\cite{DBLP:conf/aaai/BartzYM18} present a solution that employs a STN~\cite{jaderberg2015spatial} to attend to each word in the input image circularly and then recognize them individually. Li et al.~\cite{Li_2017_ICCV} substitute the object classification module in Faster-RCNN~\cite{2017Faster} with an encoder-decoder-based text recognition model and to create their text spotting system~\cite{2017Deep,he2018end,2018FOTS} develop unified text detection and recognition systems with very similar overall architectures, which consist of a recognition branch and a detection branch. Liu et al.~\cite{2020ABCNet} design a novel BezierAlign layer for extracting accurate convolution features of a text instance with arbitrary shapes and adaptively fit arbitrarily-shaped text via a parameterized Bezier curve. Liao et al.~\cite{liao2020mask} propose Mask Text Spotter v3, an end-to-end trainable scene text spotter that adopts a Segmentation Proposal Network (SPN) instead of an RPN~\cite{2017Faster}.

\subsection{Reinforcement Learning}
In earlier work, Mnih et al.~\cite{2013Playing} present the first deep learning model to successfully learn control policies directly from high-dimensional sensory input using reinforcement learning. In recent years, reinforcement learning~\cite{2017Collaborative,jie2016tree,Caicedo_2015_ICCV,mathe2016reinforcement} has evolved considerably in the field of object detection. Some works~\cite{peng2021rlst,wang2018focus} employ reinforcement learning as a post-processing method for scene text detection to adjust the bounding boxes predicted by the detection model, which can result in a significant time increase. In contrast to earlier research, we employ text recognition as the reward rather than the IOU between the predicted and ground-truth bounding boxes. In addition, our method is not a post-processing of the detection and does not add any extra computational costs to the inference phase.

\begin{figure*}
\begin{center}
\includegraphics[width=1.0\linewidth] {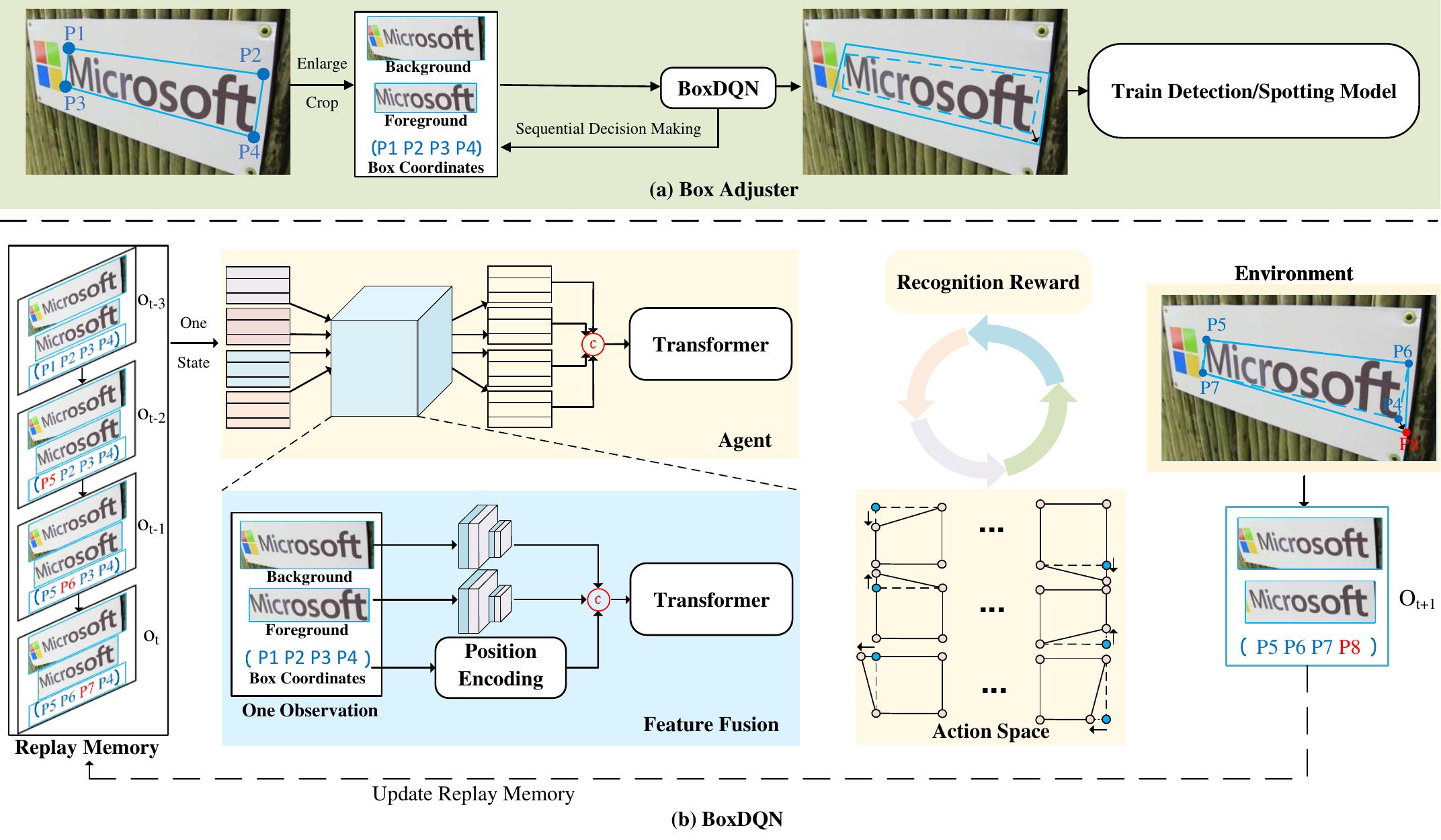}
\end{center}
  \caption{Overview of our proposed method Box Adjuster and the details of BoxDQN model architecture. In order to mitigate the inconsistency between bounding boxes and deep representations of text recognition, we utilize Box Adjuster to adjust ground-truth bounding boxes and train the text spotter with them. BoxDQN is a method based on reinforcement learning with the reward of recognition confidence.
}
\label{fig-pipeline}
\end{figure*}


\section{Methodology}

This paper aims to mitigate the inconsistency problem between bounding boxes and deep representations of text recognition. A reasonable solution is to train the detection module with suitable bounding boxes that can boost the performance of the recognition module. Thus, the issue is how to obtain these appropriate bounding boxes. As illustrated in Fig.\ref{fig-pipeline}(a), we propose a method with the BoxDQN model structure termed Box Adjuster for adjusting bounding boxes to obtain suitable shapes. BoxDQN accepts an initial bounding box and adjusts it continuously throughout the loop. Then we train the text spotter with adjusted annotated bounding boxes.

The bounding box adjustment is formulated as a sequential decision-making process. In the decision-making process, the agent constantly interacts with the environment and takes a sequence of actions to adjust the bounding box. As shown in Fig.\ref{fig-pipeline}(b), the agent chooses which action from action space to perform based on the input of four consecutive observations. Following the environment's execution of the selected action, the agent receives the next state and current reward, which can be used to guide the agent's action policy until it achieves a reasonable bounding box by maximising the cumulative rewards. In this section, we first introduce the state, action space, and reward of our model, then describe the components of BoxDQN and its training process. Finally, we detail how our method can be applied to cross-domain problems.

\subsection{State and Action Space}
\label{sec:State}
Based on the current state and reward, the agent chooses which action to take from action space. So it is crucial to capture abundant information from the state. However, one observation can only provide limited information for the agent, and it is necessary to make full use of historical observations for making decisions. Thus, we choose four serial observations as the state and the current state can be defined as $s_t$=\{$o_{t-3}$, $o_{t-2}$, $o_{t-1}$, $o_{t}$\}, where $o_{t}$ denotes the current observation at step t. A single observation is composed of background, foreground, and $\operatorname{box-coordinates}$, denoted by $o_t$=\{\textit{background}, $foreground_t$, $\operatorname{\textit{box}-\textit{coordinates}}_t$\}. The background area is four times the size of the initial bounding box. The $foreground_t$ is cropped from the background by a minimum enclosing rectangle of the bounding box at step t. The $\operatorname{\textit{box}-\textit{coordinates}}_t$ represents the coordinates of text in background at step t. We have $16$ actions in action space which are combinations of 4 vertexes and $4$ directions. As we can see from Fig.\ref{fig-pipeline}(b), the first action in action spaces implies that the top-left vertex of the quadrangle moves down by one pixel.

\subsection{Text Recognition-based Reward}
\label{sec:Reward}
The goal of BoxDQN is to capture appropriate bounding boxes for better recognition. Therefore, a reliable reward is needed to guide the agent to automatically adjust the bounding boxes. We select a few representative text recognition algorithms, including CRNN~\cite{crnn}, RARE~\cite{rare} and others. The average recognition confidence among them is regarded as a reward, so the reward at step t can be formulated as $r_t = conf_{t+1} - conf_t$, where $conf_{t}$ = $Conf(foreground_{t})$, $conf_{t}$ denotes the recognition confidence at step t.

\begin{equation}
 conf_{t} = {\sum_{k=0}^{N_P}conf_k}/max(N_G,N_P) ,\\
\end{equation}
where $N_{P}$ denotes the number of characters in a prediction word and $N_{G}$ denotes the number of characters in a ground-truth word. The aim of reinforcement learning is to maximize the cumulative rewards:
\begin{equation}
 G_t = \sum_{k=0}^{T}\gamma^{k}r_{t+k} ,\\
\end{equation}
where $\gamma$ denotes the discount factor and $\gamma$ $\in$ [0,1]. 
Ignoring the discount factor, the cumulative reward is
equal to $conf_T - conf_0$, where $conf_T$ refers to the recognition confidence of foreground in the terminal state, T means the maximum number of steps and $conf_0$ refers to the recognition confidence of foreground in the initial state. Because $conf_0$ is invariant and only determined by the initial bounding box, maximizing cumulative rewards means maximizing $conf_T$ without $\gamma$.

\subsection{BoxDQN Model}
\label{sec:BoxDQN}
With the defined action space, state, and reward, the details of BoxDQN are illustrated in Fig.\ref{fig-pipeline}(b). The agent is composed of a feature fusion module (FFM) and a transformer encoder~\cite{NIPS2017_3f5ee243}. It accepts a single state with four observations as input and outputs 16 dimensional vectors, each of which specifies the appropriate action to take. With two deep convolutional neural networks~\cite{NIPS2012_c399862d} and a transformer encoder, the FFM is proposed to integrate background, foreground, and box-coordinates. 
During a bounding box adjustment, BoxDQN receives four observations and outputs the corresponding action according to the current state. Every observation needs to be fused by the FFM successively. The feature maps of the background and foreground are extracted from two convolution neural networks, respectively. We concatenate two image feature maps and the position encoding as the input of the transformer in the FFM. After all four observations are passed through the FFM, the transformer in the agent selects an action from the action space based on the concatenation of four fused feature maps. 
The bounding box moves in response to the selected action, changing both the $\operatorname{box-coordinates}$ and the minimum enclosing rectangle of the $\operatorname{box-coordinates}$, and then the next state starts.

\begin{figure}[t]
\includegraphics[width=0.8\linewidth]{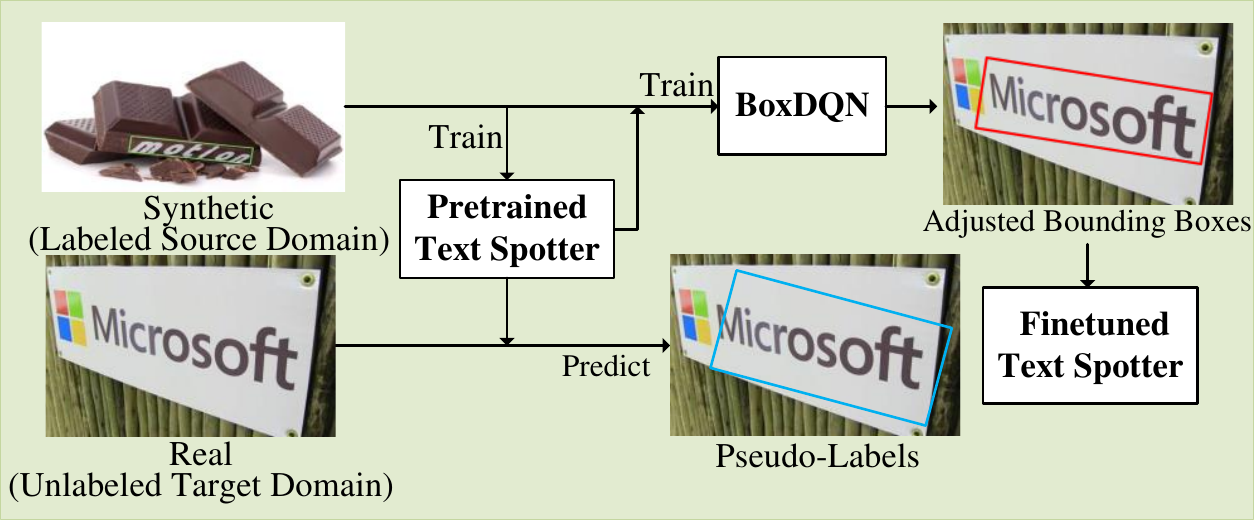}
  \centering
   \caption{Illustration of the pipeline with BoxDQN in the area of OCR domain adaptation. We propose a solution that utilizes BoxDQN to tackle domain shift problems.}
   \label{fig-domain-shift}
\end{figure}

\subsection{Domain Adaptation}
\label{sec:Domain}
In many cases, due to the absence of labeled real data, we train and test models using synthetic data. However, the domain gap between synthetic and real data degrades performance on real data. To address domain shift problems, we propose a domain-adaptive approach based on our BoxDQN. As shown in Fig.\ref{fig-domain-shift}, our method consists of four steps: (1) refer to the labeled synthetic data domain as the source domain and the unlabeled real data domain as the target domain, (2) train a text spotter with the labeled synthetic data, (3) use the trained text spotter to generate pseudo-labels on real data and adjust the pseudo-labels by employing the BoxDQN mentioned above, (4) finetune the text spotter with the adjusted bounding boxes on real data.

\subsection{Training BoxDQN Model}
\label{sec:Training}
We use a value-based reinforcement learning method to adjust the bounding boxes, and the training process is presented in Algo.\ref{training_BoxDQN}. During the inner loop of the algorithm, BoxDQN can only adjust one bounding box in a single iteration. Thus, we crop all backgrounds from the source images by the bounding boxes in advance. M is the number of backgrounds. Firstly, the agent selects and executes an action according to an $\operatorname{\epsilon-greedy}$ policy. The $\epsilon$ gradually decreases with iterations from 1.0 to 0.2. Secondly, we present two methods to determine whether the BoxDQN has reached the terminal state. Thirdly, we store a transition \{$s_t$, $a_t$, $r_t$, $s_{t+1}$, $Term_{t+1}$\} and sample random mini-batch of transitions in replay memory D. The $a_t$ represents the action at step t and $Term_{t+1}$ represents the terminal state at step t+1, respectively. $Term_{t+1}$ has two types: 0 and 1, where 0 and 1 represent termination and continuance, respectively. Finally, we refer to the training method in the paper\cite{mnih2015humanlevel} that uses a separate network termed $\hat{Q}$ for generating the targets $y_j$ in the $Q$-learning update. $Q$ represents the BoxDQN agent and has the same network structures as $\hat{Q}$. The $\operatorname{\hat{Q}-network}$ parameters $\theta^{-}$ are only updated with the $Q$-network parameters $\theta$ every C steps and are held fixed between individual updates. The parameters of $Q$ are updated by optimizing the loss function with stochastic gradient descent. The training loss function is defined as follows:
\begin{align}
  loss = (y_j - Q(s_j,a_j;\theta))^2,
\end{align}
where $y_j$ can be formulated as follows:
\begin{align}
y_j &= r_j+(1 - Term_{j+1}) * \gamma*max_{a'}\hat{Q}(a', s_{j+1}; \theta^{-}).
\end{align}

\begin{algorithm}[H]
     \caption{Training procedure of the BoxDQN Model}
     \label{training_BoxDQN}
     Initialize replay memory D to capacity N\\
     Initialize history memory H to capacity 4\\
     Initialize action-value function Q with random weight $\theta$\\
     Initialize target action-value function $\hat{Q}$ with weight $\theta^{-}=\theta$\\
     \For{$episode= 1, M$}{
         Initialize observation $o_0$ according to the initial environment $o_0=\{{background}, {foreground_0}, \operatorname{\textit{box}-\textit{coordinates}}_0\}$\\
          Store $o_0$ in H four times\\
          Initialize confidence $conf_0=Conf(foreground_0)$\\
          \For{$t= 1, T$}{
             With probability $\epsilon$ select a random action $a_t$ \\
             Otherwise select $a_t = argmax_aQ(s_t, a; \theta)$\\
             Execute action $a_t$, observe reward $r_t$, new\\
             observation $o_{t+1}$ and new confidence $conf_{t+1}$\\
             \eIf{$conf_{t+1} >= 1.2*conf_0$ or t+1==T:}
             {
                 $Term_{t+1}=1$
             }
             {
                 $Term_{t+1}=0$
             }
             Get state $s_t$ from H\\
             Update H by using $o_{t+1}$\\
             Get state $s_{t+1}$ from H\\
             Store transition $(s_t, a_t, r_t, s_{t+1}, Term_{t+1})$ in D\\
             Sample random mini-batch of transitions $(s_j, a_j, r_j, s_{j+1}, Term_{j+1})$ from D\\
             Set $s_j = H(j)$ and $s_{j+1} = H(j+1)$\\
             Set $y_j=r_j + (1-Term_{j+1})*\gamma*max_{a'}\hat{Q}(a', s_{j+1};\theta^{-})$\\
             Perform a gradient descent step on $(y_j-Q(a_j,$
             $ s_j; \theta))^2$ with respect to the network paremeters $\theta$\\
             Every C steps reset $\hat{Q} = Q$\\
             \If{$Term_{t+1}==1$}
             {
                 break out
             }
          }
    }
\end{algorithm}

\section{Experiments}


\label{sec:data}
\subsection{Datasets}
To verify the effectiveness of our method for the end-to-end text spotting methods and the classic two-step methods, we perform experiments on four different datasets. Furthermore, we conduct domain-shift experiments on these datasets to show the robustness of our method in general scenarios.

\noindent \textbf{ICDAR-2013~\cite{ic13}} (IC13) is released during the ICDAR 2013
Robust Reading Competition for focused scene text detection, consisting of high-resolution images, 229 for training and 233 for testing, containing texts in English. The annotations are at word-level using rectangular boxes.

\noindent \textbf{ICDAR-2015~\cite{ic15}} (IC15) is presented for the ICDAR 2015 Robust Reading Competition. All images are annotated with word-level and quadrilateral boxes. 

\noindent \textbf{ICDAR-2019ReCTS~\cite{rects}} (ReCTS) is a newly-released large-scale
dataset that includes 20,000 training images and 5,000 testing images, covering
multiple languages, such as Chinese, English and Arabic numerals. The images in
this dataset are mainly collected in the scenes with signboards, All text lines and
characters in this dataset are annotated with bounding boxes and transcripts.

\noindent \textbf{ICDAR-2019MLT~\cite{ic19mlt}} (MLT19) is a scene text detection dataset, including Chinese, Japanese, English, Arabic, etc. The images in MLT19 are collected from a variety of scenes, and it also contains many real-scene noises. 

\noindent \textbf{SynthText-80k~\cite{SynthText} \& SynthText-MLT~\cite{ic19mlt}} are large-scale synthetic datasets, which are adopted as pretraining for our BoxDQN and text spotting models. Furthermore, due to the difference in distribution between the synthetic and real datasets, the synthetic datasets are also used in cross-domain experiments.

\begin{figure}[t]
\begin{center}
  \includegraphics[width=1.0\linewidth]{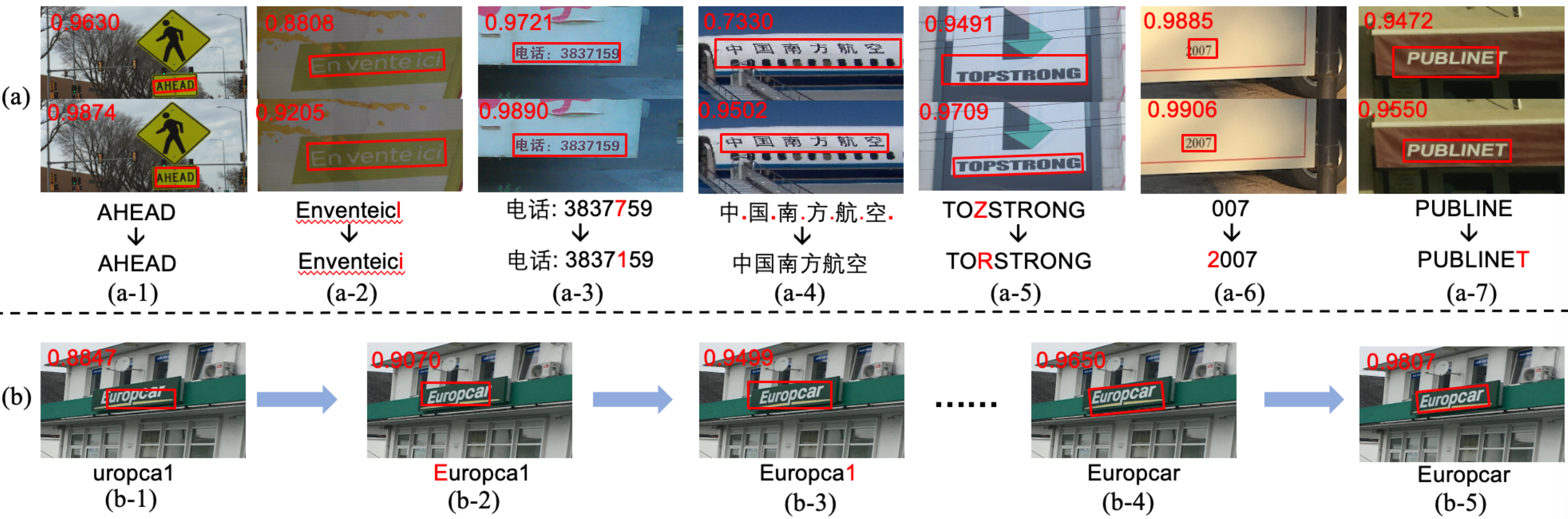}
\end{center}
   \caption{Qualitative results of BoxDQN. Each pair in (a) is a comparison between the original label (top) and our adjusted bounding boxes (bottom). (a-1) to (a-3) are the results of manual ground truth, and (a-4) to (a-7) are the results on domain-shift, whose bounding boxes are pseudo labels. (b) is a visual display of the adjustment process of BoxDQN. The upper left corner of each image uses red numbers to indicate the recognition confidence.}
\label{fig-visual}
\end{figure}

\subsection{Implementation Details}

\label{lab: Imp Detail}
\noindent \textbf{Baseline.} The baseline methods are divided into two distinct categories: (1) two-step methods and (2) end-to-end methods. For two-step methods, due to detected texts are cropped from the image, the detection and recognition are two separate steps. We choose EAST~\cite{EAST} and DBNET~\cite{DB} to detect the position of the characters, and then CRNN~\cite{crnn} and RARE~\cite{rare} recognize the content within the bounding boxes. For end-to-end methods, we adopt FOTS~\cite{2018FOTS}, ABCNET~\cite{2020ABCNet}, and MTS-V3~\cite{liao2020mask}.

\noindent \textbf{Training.} A Linux workstation with $32$ NVIDIA GeForce 2080Ti ($11$ GB) is used in our experiments. We train the recognition models in advance on SynthText-80k (for IC13 and IC15) and SynthText-MLT (for MLT19 and ReCTS) as well as on the corresponding real datasets. The trained recognition models are then used in BoxDQN training to continuously adjust and optimize the bounding boxes. The training phase of BoxDQN costs 2 days.

\noindent \textbf{Inference.} We evaluate the trained BoxDQN on the training set of the corresponding data and adjust the bounding boxes as the new ground truth to train the detection or spotting models, respectively. The average time taken for BoxDQN to adjust a bounding box is 25ms. In the process of baseline methods evaluation, we follow the official public code repository for training and testing. The datasets in the Section.\ref{sec:data} are involved in evaluation.

\subsection{Qualitative Results} 
\label{sec: Qualitative Results}

Our method mainly focuses on adjusting annotated bounding boxes for better text recognition. To verify its effectiveness, we conduct experiments on the four datasets, \ie  IC13, IC15, MLT19 and ReCTS. We show the adjustment results of our method on the bounding boxes of different datasets. The qualitative results of bounding boxes in English and Chinese are represented in Fig.\ref{fig-visual}. We find that our bounding boxes can achieve higher credibility in recognition. The (a-1) to (a-3) in Fig.\ref{fig-visual} show that our BoxDQN can adjust the bounding boxes to make them more suitable for recognition models. It can also correct inaccurate recognition of text in images.
Furthermore, we can learn from Fig.\ref{fig-visual}(b) that the adjustment steps of BoxDQN are a step-by-step process. The incorrectly labelled "Europcar" is gradually being correctly recognized. It is worth noting that our recognition confidence is increasing at each step. 
%

\subsection{Quantitative Results.} 
\label{sec: Quantitative Results}

To verify the robustness of our method with those baseline methods, we use the annotated bounding boxes refined by BoxDQN to train the baseline methods. During the quantitative evaluation, the same dataset with the original annotations is also used for training the baseline methods as a comparison. Finally, we test the F-Score metrics of our adjusted bounding boxes under recognition. The gain in Tab.\ref{Tab: two step method} indicates the gain after including our BoxDQN. 

\noindent \textbf{Two-Step Methods.} Two-step methods are those in which the detection model and the recognition model work separately. We choose the combination of [EAST, DBNET] for detection and [CRNN, RARE] for recognition in our experiments. The BoxDQN enhances the bounding boxes, and the detection models are then trained on the adjusted training part of the datasets. The trained models are then evaluated on the test datasets, respectively. From Tab.\ref{Tab: two step method}, when any combination of two-step pipelines is trained on our BoxDQN refined data, the metrics obtained are greatly improved. We find that our method has a greater improvement for MLT19 in Tab.\ref{Tab: two step method}. The gain of [EAST+CRNN] on the IC15 is $1.7\%$, but for the same pipeline on MLT19, the gain is $3.0\%$. For more complex OCR scenarios, our method has a more significant improvement. Regardless of any pipeline, the gain of the F-Score can obtain an improvement of at least $1.6\%$, which is robust to two-step methods.

\noindent \textbf{End-to-End Methods.} There are some differences between the end-to-end methods and the two-step methods, mainly in the independence of the detection branch and the recognition branch.
We adopt the bounding boxes adjusted by BoxDQN to train the whole end-to-end models rather than the detection models and test the appearance of each metric. Tab.\ref{Tab: e2e method} shows the results of different end-to-end methods. Our BoxDQN is also helpful for the end-to-end text spotting methods, especially for ABCNET, whose gain is $3.3\%$ and $2.6\%$ on the MLT19 and ReCTS, respectively. The gain of the end-to-end methods is slightly lower compared with the two-step methods. This may be due to the fact that the end-to-end training of text spotters can slightly mitigate inconsistencies in detection and recognition.

\begin{table}[t]
\footnotesize
\caption{The quantitative results of our method on the two-step methods. Gain stands for the improvement of the F-Score with and without BoxDQN. We bold the results of each gain to highlight the improvement of the effect by BoxDQN.}
  \begin{center}
  \setlength\tabcolsep{2pt}
  \begin{tabular}{l |c|c|c|c|c|c|c|c|c}
  \hlinew{1.1pt}
  \multirow{2}{*}{\textbf{Methods}} &\multirow{2}{*}{\textbf{BoxDQN}}& \multicolumn{2}{c|}{\textbf{IC13}} & \multicolumn{2}{c|}{\textbf{IC15}} & \multicolumn{2}{c|}{\textbf{MLT19}}  & \multicolumn{2}{c}{\textbf{ReCTS}}\\
  \cline{3-10}
   & &F-score &Gain &F-score &Gain &F-score &Gain &F-score &Gain\\
   \hline
  \multirow{2}{*}{\scriptsize{[EAST+CRNN]}} & \textbf{--} &  84.7 &  \multirow{2}{*}{\textbf{2.1}} &  82.2 &  \multirow{2}{*}{\textbf{1.7}} &  53.9 &  \multirow{2}{*}{\textbf{3.0}} &  69.7 &  \multirow{2}{*}{\textbf{2.8}} \\
  & \checkmark &  86.8 & &  83.9 & &  56.9 & & 72.5  \\
  \hline
  \multirow{2}{*}{\scriptsize{[EAST+RARE]}} & \textbf{--} & 85.5 & \multirow{2}{*}{\textbf{1.8}} &  83.7 &  \multirow{2}{*}{\textbf{1.9}} &  55.5 &  \multirow{2}{*}{\textbf{2.9}} & 71.1 &  \multirow{2}{*}{\textbf{2.6}} \\
  & \checkmark &  87.3 &   &  85.4  &   &  58.1 &   &  73.7  \\
  \hline
  \multirow{2}{*}{\scriptsize{[DBNET+CRNN]}} & \textbf{--} &  85.2 &  \multirow{2}{*}{\textbf{1.9}} &  83.4 &  \multirow{2}{*}{\textbf{1.7}} &  55.9 &  \multirow{2}{*}{\textbf{2.7}} &  70.0 &  \multirow{2}{*}{\textbf{2.9}} \\
  & \checkmark &  87.1 &   &  85.1 &   &  58.6 &   &  72.9  \\
  \hline
  \multirow{2}{*}{\scriptsize{[DBNET+RARE]}} & \textbf{--} &  85.4 &  \multirow{2}{*}{\textbf{1.8}} &  84.7 &  \multirow{2}{*}{\textbf{1.6}} &  57.2 &  \multirow{2}{*}{\textbf{2.4}} &  73.4 &  \multirow{2}{*}{\textbf{2.1}}\\
  & \checkmark &  87.2 &   &  86.3 &   &  59.6 &   &  75.5    \\
  \hlinew{1.1pt}
  \end{tabular}
  \end{center}
\label{Tab: two step method}
\end{table}

\begin{table}[t]
\footnotesize
\caption{The quantitative results of our method on the end-to-end methods. The metrics are the same as the Tab.\ref{Tab: two step method}.}
  \begin{center}
  \setlength\tabcolsep{2pt}
  \begin{tabular}{l |c|c|c|c|c|c|c|c|c}
  \hlinew{1.1pt}
  \multirow{2}{*}{\textbf{Methods}} &\multirow{2}{*}{\textbf{BoxDQN}} & \multicolumn{2}{c|}{\textbf{IC13}~\cite{ic13}} & \multicolumn{2}{c|}{\textbf{IC15}~\cite{ic15}}& \multicolumn{2}{c|}{\textbf{MLT19}~\cite{ic19mlt}}& \multicolumn{2}{c}{\textbf{ReCTS}~\cite{rects}}  \\
  \cline{3-10}
   &  & F-Score & Gain  & F-Score & Gain & F-Score & Gain & F-Score & Gain \\
   \hline
  \multirow{2}{*}{FOTS~\cite{2018FOTS}} & \textbf{--} &83.7 &\multirow{2}{*}{\textbf{1.9}} &81.5 &\multirow{2}{*}{\textbf{1.8}} &53.0 &\multirow{2}{*}{\textbf{3.1}} &70.2 &\multirow{2}{*}{\textbf{2.7}} \\
  & \checkmark &  85.6 & & 83.3 &  & 56.1 & &  72.9 &   \\
  \hline
  \multirow{2}{*}{ABCNET~\cite{2020ABCNet}} & \textbf{--} &86.8 &\multirow{2}{*}{\textbf{1.6}}   &82.4 &\multirow{2}{*}{\textbf{1.7}} &56.2 &\multirow{2}{*}{\textbf{3.3}}&72.5 &\multirow{2}{*}{\textbf{2.6}} \\
  & \checkmark &  88.4 &   &  84.1&  &59.5 &   &  75.1 &  \\
  \hline
  \multirow{2}{*}{MTS-V3~\cite{liao2020mask}} & \textbf{--} &  87.6 &  \multirow{2}{*}{\textbf{1.5}} &  83.1 &  \multirow{2}{*}{\textbf{1.4}} & 61.2 &\multirow{2}{*}{\textbf{2.8}} &73.4 &\multirow{2}{*}{\textbf{2.3}} \\
  & \checkmark & 89.1 &   &84.5& &  64.0 &  &75.7 &   \\
  \hlinew{1.1pt}
  \end{tabular}
  \end{center}
\label{Tab: e2e method}
\end{table}

\begin{table}[t]
\caption{The experiments results on the cross-domain unlabeled datasets, the annotation information do not used in the datasets. We bold the gain of each pipelines.}
\small
  \begin{center}
  \setlength\tabcolsep{2pt}
  \begin{tabular}{l |c|c|c|c|c}
  \hlinew{1.1pt}
  \multirow{2}{*}{\textbf{Methods}} &\multirow{2}{*}{\textbf{BoxDQN}} & \multicolumn{2}{c|}{\textbf{IC15}~\cite{ic15}} & \multicolumn{2}{c}{\textbf{MLT19}~\cite{ic19mlt}}  \\
  \cline{3-6}
   & &F-Score &Gain &F-Score &Gain \\
   \hline
  \multirow{2}{*}{[EAST~\cite{EAST}+CRNN~\cite{crnn}]} & \textbf{--} &  67.1  &  \multirow{2}{*}{\textbf{5.4}} &  41.3  &  \multirow{2}{*}{\textbf{7.0 }}  \\
  & \checkmark &  72.5  &    &  48.3   \\
  \hline
  \multirow{2}{*}{[EAST~\cite{EAST}+RARE~\cite{rare}]} & \textbf{--} & 68.3 &  \multirow{2}{*}{\textbf{5.3}} & 42.9  &  \multirow{2}{*}{\textbf{6.7}} \\
  & \checkmark & 73.6  &    & 49.6   \\
  \hline
  \multirow{2}{*}{[DBNET~\cite{DB}+CRNN~\cite{crnn}]} & \textbf{--} & 68.3 &  \multirow{2}{*}{\textbf{5.2}}  & 42.7 &  \multirow{2}{*}{\textbf{5.4}} \\
  & \checkmark & 73.5 &  & 48.1   \\
  \hline
  \multirow{2}{*}{[DBNET~\cite{DB}+RARE~\cite{rare}]} & \textbf{--} & 69.4 &  \multirow{2}{*}{\textbf{5.3}} &  44.3  &  \multirow{2}{*}{\textbf{6.2}} \\
  & \checkmark  & 74.7 &   & 50.5   \\
  \hline
  \hline
  \multirow{2}{*}{[FOTS~\cite{2018FOTS}]} & \textbf{--} & 66.6 &  \multirow{2}{*}{\textbf{4.8}} & 39.8 &  \multirow{2}{*}{\textbf{5.8}} \\
  & \checkmark & 71.4 &   & 45.6   \\
  \hline
  \multirow{2}{*}{[ABCNET~\cite{2020ABCNet}]} & \textbf{--} & 69.4 &  \multirow{2}{*}{\textbf{4.4}} & 44.4 &  \multirow{2}{*}{\textbf{5.1}} \\
  & \checkmark & 73.8 &   & 49.5   \\
  \hline
  \multirow{2}{*}{[MTS-V3~\cite{liao2020mask}]} & \textbf{--}  & 71.3 &  \multirow{2}{*}{\textbf{4.7}} & 46.7 &  \multirow{2}{*}{\textbf{4.6}} \\
  & \checkmark & 76.0 &   & 51.3   \\
  \hlinew{1.1pt}
  \end{tabular}
  \end{center}
\label{Tab: domain adaption}
\end{table}

\subsection{Domain Adaption}
\label{Domain Adaption}
Taking into account the domain gap between the synthetic pretraining datasets and the in-the-wild data, we conduct cross-domain experiments to verify the generalization of our method. In detail, we pretrain BoxDQN on the synthetic datasets (SynthText-80k~\cite{SynthText} and Synthetic-MLT~\cite{ic19mlt}). After that,  we adopt the pre-trained detection models on the relevant real datasets to obtain pseudo bounding boxes. The BoxDQN adjusts the pseudo bounding boxes, and finally the recognition models work on the adjusted bounding boxes to obtain the recognition results. We simulate in-the-wild data through unlabeled IC15 and MLT19, and verify the domain adaptability of our BoxDQN. The results of domain adaption experiments are shown in Tab.\ref{Tab: domain adaption} and Fig.\ref{fig-visual}(a-1,2,3). From Fig.\ref{fig-visual}(a-4,5,6,7), although our method has a slight visual deviation in the adjustment of pseudo-labels, it can improve the confidence and correct the wrong recognition results. Tab.\ref{Tab: domain adaption} proves that our method can improve at least $4.4\%$ in cross-domain datasets. 

\begin{table}[t]
\footnotesize
\caption{Ablation study on grid search. The effect comparison under our BoxDQN and the grid search policy. The experimental dataset is based on IC15~\cite{ic15}.}
  \begin{center}
  \setlength\tabcolsep{2.3pt}
    \begin{tabular}{c|ccc|ccc}
    \hlinew{1.1pt}
    \multirow{2}{*}{Methods}  & \multicolumn{3}{c|}{Grid Search} & \multicolumn{3}{c}{BoxDQN}  \\
      \cline{2-7} 
      & Precision & Recall & F-Score & Precision & Recall & F-Score \\
      \hline
      $[$EAST~\cite{EAST}+CRNN~\cite{crnn}$]$ & 90.3 & 76.4 & 82.8 & 91.0 & 77.8 & \textbf{83.9} \\
      $[$DBNET~\cite{DB}+RARE~\cite{rare}$]$ & 91.9 & 80.3 & 85.7 & 92.5 & 80.8 & \textbf{86.3}  \\
      $[$ABCNET~\cite{2020ABCNet}$]$ & 93.6 & 74.5 & 83.0 & 94.6 & 75.7 & \textbf{84.1} \\
      $[$MTS-V3~\cite{liao2020mask}$]$ & 93.5 & 75.2 & 83.4 & 94.8 & 76.2 & \textbf{84.5}  \\
      \hlinew{1.1pt}
    \end{tabular}
  \end{center}
\label{table_ab_study_1}
\end{table}

\subsection{Ablation Study}
\noindent \textbf{Grid Search.} 
To verify that our BoxDQN is reasonable for adjusting annotated bounding boxes, we compare our method with a grid search policy and include the metric of F-Score in this experiments. In detail, we perform a grid search in each bounding box's four vertices in the directions of up, down, left, and right with a step length of one pixel. The recognition models are trained in advance and give the results with the highest confidence after $10$ rounds of grid search as the new ground truth. Refer to Tab.\ref{table_ab_study_1} for the quantitative results. When compared with grid search, BoxDQN can improve recognition accuracy. This qualifies it as an appropriate bounding-box adjustment method in OCR systems and indicates that it does not over-fit the datasets.

\begin{table}[t]
\small
\caption{Ablation study on the DQN with only foreground image as input. The effect between our BoxDQN and the original DQN is shown below. The dataset is IC15~\cite{ic15}.}
  \begin{center}
  \setlength\tabcolsep{2.3pt}
    \begin{tabular}{c|ccc|ccc}
    \hlinew{1.1pt}
    \multirow{2}{*}{Methods}  & \multicolumn{3}{c|}{DQN} & \multicolumn{3}{c}{BoxDQN}  \\
      \cline{2-7} 
      & Precision & Recall & F-Score & Precision & Recall & F-Score \\
      \hline
      $[$EAST~\cite{EAST}+CRNN~\cite{crnn}$]$ & 90.5 & 77.2 & 83.3 & 91.0 & 77.8 & \textbf{83.9} \\
      $[$DBNET~\cite{DB}+RARE~\cite{rare}$]$ & 92.3 & 79.9 & 85.7 & 92.5 & 80.8 & \textbf{86.3}  \\
      $[$ABCNET~\cite{2020ABCNet}$]$ & 94.0 & 74.5 & 83.1 & 94.6 & 75.7 & \textbf{84.1} \\
      $[$MTS-V3~\cite{liao2020mask}$]$ & 94.1 & 75.5 & 83.8 & 94.8 & 76.2 & \textbf{84.5}  \\
      \hlinew{1.1pt}
    \end{tabular}
  \end{center}
\label{table_ab_study_2}
\end{table}
%

\noindent \textbf{Only Foreground Image as Input.} Our BoxDQN model adopts a FFM that fuses the foreground, background, and coordinates from the text images. The experimental settings in this section are the same as those in the Sec.\ref{lab: Imp Detail}. The comparison results are shown in Tab.\ref{table_ab_study_2}, demonstrating that the BoxDQN with more prior information outperforms the classic DQN ($86.3$ vs. $85.7$,  [DBNET~\cite{DB}+RARE~\cite{rare}] row). And for all of the representative methods, our method has a steady improvement on them. This verifies the robustness of our BoxDQN. More importantly, with the background image as input, our model can handle cases such as those shown in Fig.\ref{fig-visual}(a-7) where the bounding box is slightly shorter than the text transcription.

\begin{table}[t]
\footnotesize
\caption{Ablation study on the number of iterations of BoxDQN. Each row shows the F-Score of the BoxDQN with a different iteration number. We bold the best value of each column.}
  \begin{center}
  \setlength\tabcolsep{2pt}
    \begin{tabular}{l |ccccc}
    \hlinew{1.1pt}
      \multirow{1}{*}{Iter} &[EAST+CRNN] &[DBNET+RARE] & [FOTS] & [ABCNET] & [MST-V3]   \\
      \hline
      { 5}          & 82.9 & 85.2 & 82.3 & 82.9 & 83.5 \\
      {10}          & 83.3 & 85.8 & 82.7 & 83.4 & 84.0  \\
      {20}          & 83.9 & \textbf{86.3} & \textbf{83.3} & 84.1 & 84.5  \\
      {40}          & \textbf{84.0} & 86.2 & 83.1 & \textbf{84.3} & \textbf{84.6}  \\
      \hlinew{1.1pt}
    \end{tabular}
  \end{center}
\label{tab_ab4}
\end{table}

\noindent \textbf{BoxDQN under Different Iterations.} Since our BoxDQN is sensitive to the times of iterations, different iterations have a great impact on the effect. We test the BoxDQN under different iterations and compare the number of iterations corresponding to the best BoxDQN. As shown in Tab.\ref{tab_ab4}, the best performance of BoxDQN can be achieved when the number of iterations is set at $20$, with minimal resource consumption. This is the same as the number of iterations (20) set in our experiments.




\subsection{Exploration on Arbitrarily-shaped Text based on Bezier Curves} 
To further explore the potential of our approach, we perform experiments on arbitrarily shaped text (TotalText~\cite{totaltext} dataset).
Our BoxDQN method requires a text representation with a fixed number of boundary points for optimization, so we have to convert polygon contour points that do not have a fixed number of points to a representation that does. We can currently only convert arbitrarily-shaped text to a fixed number of control points ($8$) with the help of Bezier curves.  We train our BoxDQN on the SynText150k~\cite{2020ABCNet} dataset which contains 150k synthetic arbitrary-shaped text annotated by Bezier curves. Then, we use BoxDQN to adjust the control points of the Bezier curve to obtain the optimal ground truth. The rest of the experimental settings is consistent with the multi-oriented text datasets, except for the differences mentioned above. From Tab.\ref{Tab: totaltext}, we can find there is a considerable improvement($61.5$ \textit{vs.} $63.8$) in recognition performance when training with the ground-truth Bezier curves optimized by BoxDQN. This experiment illustrates the possibility of extending our approach to arbitrarily-shaped text if there is a more general representation of text boxes. 

\begin{table}[t]
\footnotesize
\caption{The quantitative results of our method on TotalText. The baseline model is ABCNet with Bezier curves.}
  \begin{center}
  \setlength\tabcolsep{2pt}
  \begin{tabular}{l |c|c|c}
  \hlinew{1.0pt}
  \multirow{2}{*}{\textbf{Methods}} &\multirow{2}{*}{\textbf{BoxDQN}}& \multicolumn{2}{c}{\textbf{TotalText}~\cite{totaltext}}\\
  \cline{3-4}
   && F-Score & Gain \\
    \hline
  \multirow{2}{*}{ABCNet~\cite{EAST}} & \textbf{--} &61.5 &\multirow{2}{*}{\textbf{2.3}}\\
  & \checkmark &  63.8 & \\
 
  \hlinew{1.0pt}
  \end{tabular}
  \end{center}
\label{Tab: totaltext}
\end{table}


\section{Conclusion and Future Work}

\label{sec:conclusion}
In this work, we first analyze the inconsistency between bounding boxes and text recognition, and then present a novel and general preprocessing method called Box Adjuster, which learns the optimal distribution of the text recognition module and delivers it to detection via bounding box adjustment. Our proposed approach is employed exclusively during the training phase, with no additional calculations during the prediction phase. More significantly, the cross-domain problems will be alleviated by utilizing the Box Adjuster. 
Comprehensive experiments have demonstrated that the proposed approach rationally addresses the aforementioned inconsistency and significantly improves the performance of both two-step and end-to-end text spotting approaches on standard datasets. 
In the future, we hope to extend our method for arbitrary-shaped text spotting.

\section{Acknowledgements}
\label{sec:ack}
This work  was supported by the National Natural Science Foundation of
China 61733007.

\clearpage
%
%
\bibliographystyle{splncs04}
\bibliography{egbib}
\end{document}